\documentclass[conference]{IEEEtran}
\usepackage[latin9]{inputenc}
\usepackage{amsmath}
\usepackage{graphicx}
\usepackage[official]{eurosym}
\usepackage{pbox}

\usepackage{cite}

\ifCLASSINFOpdf
\else
\fi

\usepackage{amsmath}
\usepackage{amssymb}

\usepackage{array}
\usepackage[caption=false,font=footnotesize]{subfig}
\usepackage{stfloats}

\usepackage{url}
\PassOptionsToPackage{hyphens}{url}
\usepackage{hyperref}

\DeclareMathOperator*{\argmax}{arg\,max}
\def\argdot{{\hspace{0.18em}\cdot\hspace{0.18em}}}

\begin{document}
%
\title{A feasibility study of deep neural networks for the recognition of banknotes regarding central bank requirements}

\author{\IEEEauthorblockN{Julia Schulte }
\IEEEauthorblockA{CI Tech Sensors AG \\
Email: julia.schulte@citechsensors.com}
\and
\IEEEauthorblockN{Daniel Staps}
\IEEEauthorblockA{ Mittweida University\\ of Applied Science\\
Email:  dstaps@hs-mittweida.de}
\and
\IEEEauthorblockN{Alexander Lampe}
\IEEEauthorblockA{Mittweida University\\ of Applied Science\\
Email:  lampe@hs-mittweida.de }}


%

\maketitle

\begin{figure}[b]
\vspace{-0.3cm}
\parbox{\hsize}{\em
}\end{figure}

\begin{abstract}
This paper contains a feasibility study of deep
neural networks for the classification of Euro banknotes with
respect to requirements of central banks on the ATM and high speed sorting
industry. 
Instead of concentrating on the accuracy for a large
number of classes as in the famous ImageNet Challenge we
focus thus on conditions with few classes and the requirement of
rejection of images belonging clearly to neither of the trained classes (i.e.
classification in a so-called 0-class). These special requirements
are part of frameworks defined by central banks as the European Central
Bank and are met by current ATMs and high speed sorting
machines. 
We also consider training and classification time on
state of the art GPU hardware. The study concentrates on the banknote recognition whereas banknote class dependent authenticity and fitness checks are a topic of its own which is not considered in this work.
\end{abstract}


%
\IEEEpeerreviewmaketitle

\section{Introduction}

An important step in the recirculation of banknotes is their recognition in high-speed sorting machines and cash-recycling ATMs which is followed by banknote class dependent authenticity and fitness checks. Central banks ensure the integrity of the cash cycle by formulating requirements and testing relevant banknote handling machines see e.g. \cite{ECB_Decision_2012}.
On the other hand in the last decade deep learning has outperformed classical algorithms in disciplines like natural language processing, speech recognition or image recognition see \cite{Goodfellow-et-al-2016}. 
 
In the current paper we use a deep learning approach for the recognition of Euro banknotes which seems to be a simple challenge at first glance but in fact it turns out to be tricky when central bank requirements with respect to rejection of objects not belonging to a trained banknote class are taken into consideration. 

The success of deep learning is mainly driven by challenges like the ImageNet Large Scale Visual Recognition Challenge (ILSVRC) with 1000 classes including keyboard, mouse, pencil, and many animals and more than one million images, see \cite{russakovsky2015imagenet}.
A milestone in the history of deep learning was the success of Alexnet \cite{krizhevsky2012imagenet}, the first convolutional neural network (CNN), winning the ILSVRC challenge by reduction of the top-5 error rate from 26.1\% to 15.3\%.
 The achievements of deep learning are made possible by advances in available computing power and larger training data sets allowing for deeper models with millions of parameters.
Following the popularity of deep learning in academia it is now available to the industry and general public via open source frameworks as e.g. Tensorflow \cite{abadi2016tensorflow} or Matlab's deep learning toolbox. Since in most applications massive amounts of training data are not available a popular approach consists of transfer learning, i.e. the finetuning of some layers of the pretrained network for a specific task, where the pretraining is done on a similar but sufficiently large data set, see \cite{yosinski2014transferable,donahue2014decaf}. An example is the pretraining on Imagenet and the fine tuning on a smaller data set for the classification of plants \cite{reyes2015fine} or medical images \cite{shin2016deep,hon2017towards}. 
For an overview of different studies dealing with banknote recognition we refer to \cite{LeeHongKimPark2017Survey} where also counterfeit banknote detection, serial number recognition and fitness classification are covered. To our knowledge the first article applying deep neural networks to the problem of banknote recognition is \cite{PhamNguyenParkPark2019DL} which uses a combination of visible-light and infrared-light images. Though, in comparison with the present work, the rejection of images with wrong image or format is not considered. 
The problem of recognition with a reject class is often referred to as open set recognition \cite{Scheirer2012}. A sophisticated probabilistic approach accompanied by an experimental study on the ImageNet database is contained in \cite{bendale2016towards}.  \\ 
The current paper is structured as follows.
In Section \ref{sec:background} we introduce the necessary background on banknote categories and the recycling framework introduced by the European Central Bank. Subsequently we summarize some background on image classification using deep neural networks and transfer learning in a mathematically rigorous way. 
In Section \ref{Sec:ProposedMethod} we introduce the 0-class module, i.e. we propose an architecture modification of the classical deep neural network image classifier which allows the mapping of images to a reject class. Then, we introduce the Inception-v3 neural network which will be applied for transfer learning.
In Section \ref{Sec:ExpStudy} we present the results of our experimental study which contains a statistical analysis of classification results for genuine banknotes under different training conditions and on 3 data sets with different resolutions and with or without application of skew correction. We study classification results for a deep neural network classifier with additional 0-class module also regarding reject rates on genuine banknotes as well as on images not belonging to a trained banknote class, where rejection is desired for the latter.  
Furthermore, we study training and test times. A summary of the results is contained in Section \ref{Sec:Con}.

\section{Background}\label{sec:background}
\subsection{Banknote categories and banknote recycling framework by the European Central Bank}

Ensuring the integrity of the banknote cycle is a major task of the European Central Bank (ECB). For this reason the recirculation as well as the authenticity and fitness checking is regulated in the decisions \cite{ECB_Decision_2010,ECB_Decision_2012} which apply to bank note handling machines like cash-recycling ATMs and high-speed sorting machines used e.g. in banks and cash-centers. The ECB distinguishes between 4 categories of inputs to banknote handling machines. Category 1 consists of objects not recognized as Euro banknotes e.g. because of a wrong image or format, transportation errors, large folded corners, missing parts or non-Euro currency, compare Figure \ref{Cat1_ExampleImages}. Category 1 objects have to be rejected to the customer or operating staff. Category 2 consists of suspect counterfeit Euro notes where image and format are correct but one ore more security features are clearly missing or out of tolerance. Category 2 shall be withdrawn from circulation, handed over to the national authorities within 20 working days for further investigation and not be credited to the account holder. Category 3 consists of banknotes with correct image and format but some security features which cannot clearly be authenticated possibly due to tolerance deviations or bad quality of the banknote. Category 3 is treated in the same way as category 2 but may be credited to the account holder. Category 4 consists of genuine banknotes where all authenticity checks are positive. Furthermore it is differed between Category 4a where also all fitness checks have a positive results and category 4b where at least some fitness criteria has a negative result. Category 4 is credited to the account holder but only category 4a shall be used for recirculation whereas category 4b shall be returned to a national central bank.
The ECB tests banknote handling machines and publishes a list of successfully tested machines on their website \cite{CertifiedBHM_CustOp_ECB2019,CertifiedBHM_StaffOp_ECB2019}. The current test procedure (see \cite{ECBTestProcedure}) contains a counterfeit test (at least 90\% of a given counterfeit test deck shall be sorted in category 2 or 3 and none in category 4) and a fitness test (not more than 5\% of a given test deck of unfit notes shall be sorted to category 4a). Furthermore at least 90\% of a given test deck of fit and genuine notes shall be classified as category 4a and not more than 1\% of the fit and genuine notes as category 1, 2 or 3.

\subsection{Image classification and deep neural networks}

An image classifier is a function 
\[
f: [a,b]^{w\times h\times c}\to \{1,\ldots,N\},
\]
where $h,w,c\in\mathbb{N}$ are the width and height of the input image in pixels, $c\in\mathbb{N}$ is the number of input channels (usually 3 for red, green and blue) and $[a,b]$ is the range of the pixel values. Furthermore $N\in\mathbb{N}$ denotes the number of image classes.
Let 
\[
PV(N)=\{(y_1,\ldots,y_N)\in[0,1]^{N} \text{ with } y_1+\ldots +y_N=1\}
\]
denote the space of probability vectors of length $N$.
A deep neural net is a function 
\[
f_{DNN}(\argdot;\Theta):[a,b]^{w\times h\times c}\to PV(N),
\]
where $\Theta\in\mathbb{R}^p$ is the vector of trainable parameters with $p>10^6$ not being unusual.
A deep neural net can be used as image classifier of the form
\[
f=g\circ f_{DNN}(\argdot;\Theta),
\]
where
\[
g(y)=\argmax(y),\quad y\in PV(N).
\]
Training a neural net means that the parameter vector $\Theta$ is repeatedly updated in so-called training episodes where a chosen stochastic optimization algorithm is used to optimize the parameters for the classification of a randomly chosen batch of training images. Batch sizes of several hundred and training episode numbers of more than 1000 are not unusual.
 
\subsection{Transfer learning}
Let a deep neural net $f_{DNN}$ mapping to $PV(N)$ be given. 
Usually $f_{DNN}$ has a decomposition of the form
\[
f_{DNN}=f_{class}(\argdot;\Theta_2)\circ f_{feat}(\argdot;\Theta_1),
\]
where 
\[
f_{feat}(\argdot;\Theta_1):[a,b]^{w\times h\times c}\to \mathbb{R}^l
\]
with trainable parameters $\Theta_1\in \mathbb{R}^{m}$ is the mapping to the so called feature space and 
\[
f_{class}(\argdot;\Theta_2):\mathbb{R}^l\to PV(N)
\]
with trainable parameters $\Theta_2\in\mathbb{R}^{k}$ consists of the last two layers of the deep neural network which are a fully connected layer followed by a softmax layer.
By concatenating the feature map $f_{feat}$ of a given deep neural net with a suitable $f_{class}$ mapping to $PV(n)$ instead of $PV(N)$ we can obtain a deep neural network for the classification of $n$ image classes instead of $N$ which the network was originally trained with.

By transfer learning we mean the training of a classifier 
\[
f=g\circ f_{class}(\argdot;\Theta_2)\circ f_{feat}(\argdot;\Theta_1)
\]
with fixed parameters $\Theta_1\in\mathbb{R}^{m}$ for the classification of $n$ different image classes by updating only the parameter vector $\Theta_2\in\mathbb{R}^k$. Usually $k$ is much smaller than $p$. Thus transfer learning has a significantly reduced training time in comparison to training from scratch.

\section{Proposed Method}\label{Sec:ProposedMethod}

\subsection{The 0-class module}
We introduce a reject class or 0-class which contains images belonging to neither of the trained classes. 
For this purpose we introduce the so called 0-class module 
\[
f_{\text{0-class}}(\argdot;T): PV(n)\to \{0,\cdots,n\}
\]
defined by  
\[
f_{\text{0-class}}(y;T)=\begin{cases}0, &\max(y)\leq T,\\ \argmax(y), &\max(y)>T, \end{cases} 
\]
for $y\in PV(n)$ with parameter $T\in[0,1]$ which can be either trained or chosen by hand.
Thus the deep neural network classifier with 0-class module is of the form
\[
f=f_{\text{0-class}}\circ f_{class}\circ f_{feat}: [a,b]^{w\times h\times c}\to \{0,\ldots,n\},
\]
where $f(x)=0$ means a reject of image $x$ and $f(x)=i$ for $i\in\{1,\ldots,n\}$ means that the image $x$ is mapped to the $i$th trained class.

\subsection{Google's inception-v3 CNN}
\begin{figure}[ht]
	\centering
		\includegraphics[width=8cm]{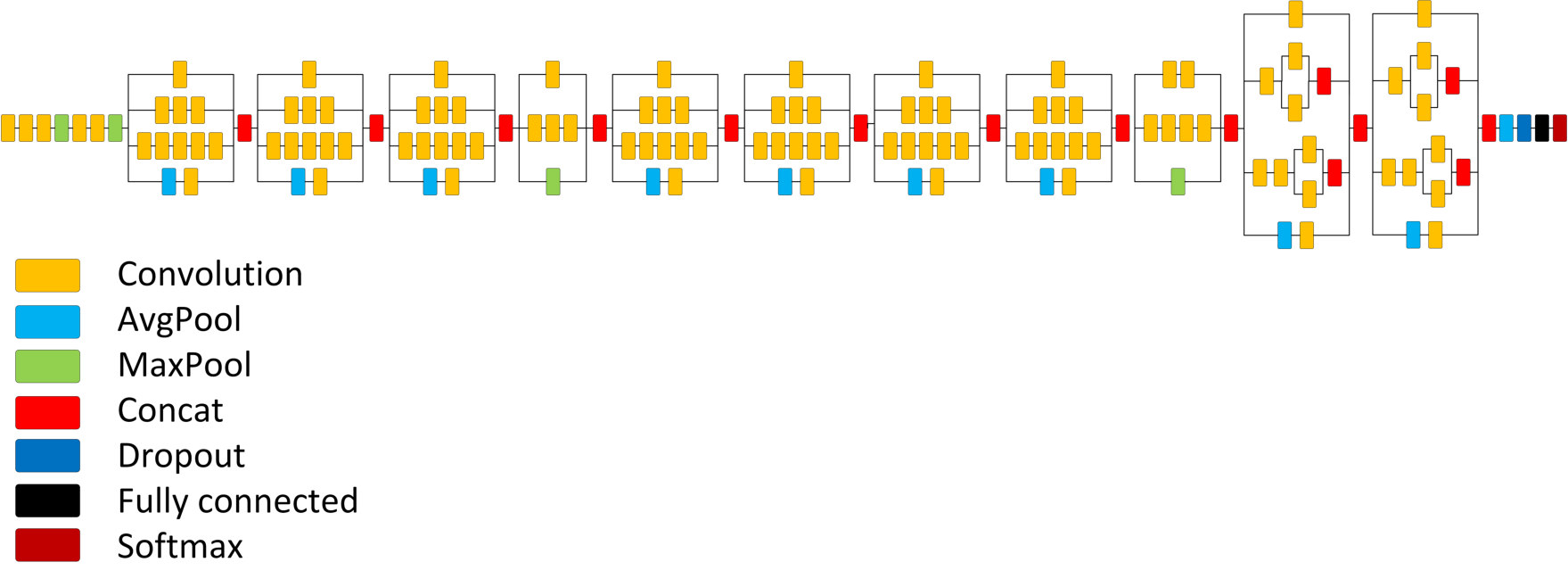}
		\caption{Architecture of the Inception-v3 neural network.}
	\label{fig:InceptionArchitecture}
\end{figure}

As deep neural net we choose in the following Google's Inception-v3 architecture proposed in \cite{szegedy2016rethinking}, which achieves a 3.45\% top-5 error rate on the ILSVRC2012 benchmark data set.
The inception-v3 has a feature map of the form 
\[
f_{feat}(\argdot;\Theta_1):[-1,1]^{299\times 299\times 3}\to \mathbb{R}^{2048},
\]
with trainable parameter vector $\Theta_1\in\mathbb{R}^m$ with $m=23885392$. 
The inception architecture consists of 48 different network layers, compare Figure \ref{fig:InceptionArchitecture}, which are mostly convolutional, pooling or inception layers and have the ability to detect edges, corners, contours and objects. It is available for transfer learning, see \cite{Inceptionv3_Pretained}, with parameters pretrained on more than one million images from the ImageNet database \cite{russakovsky2015imagenet}.
\section{Experimental Study}\label{Sec:ExpStudy}
\subsection{The hardware}
To assess the potential of the described approach for application in the ATM industry we made an experimental study on a work station with the state of the art GPU NVIDIA GeForce GTX 1080 Ti with 11 GB graphics storage and a price of 759\euro{}. The transfer learning for the inception-v3 net and our architecture modifications were done in Google's open source framework Tensorflow \cite{abadi2016tensorflow} building on the pretrained model available as open source repository, see \cite{Inceptionv3_Pretained}.
\subsection{The data sets}
\begin{figure}[ht]
\subfloat[Example image from data set 1 of class EUR\_005\_a\_1.]{ 
\includegraphics[resolution=25,scale=0.25]{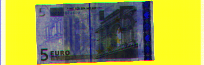}}\\
\subfloat[Example image from data set 2 of class EUR\_020\_b\_4.]{
 \includegraphics[resolution=25,scale=0.25]{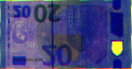}}\quad
\subfloat[Example image from data set 3 of class EUR\_500\_a\_4.]{
 \includegraphics[resolution=8,scale=0.25]{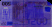}}
\caption{Example images for data set 1, 2 and 3 displayed at 25\% of the original size.}
\label{ExampleImages}
\end{figure}

We compare results for three different data sets of Euro banknotes which are provided by Diebold Nixdorf AG and consist of field data recorded with a high-speed line camera used in Diebold Nixdorf cash recycling ATMs, see \cite{DN_CS4060}, with integrated sensor module and recognition software by CI Tech Sensors AG, see \cite{CITechSensorsAG}. The first aim of the experimental study conducted in 2018 under non-disclosure agreement with Mittweida University of Applied Science was to asses the potential for industrial application. This explains why the data is not made publicly available. Still the authors promote publication of the results for the sake of public interest in a balanced view on deep learning methods. 
All data sets consist of images with three color channels with the specialty that red and green are recorded in transmitted illumination whereas blue is recorded in reflected illumination. The resolution is $25$dpi for data sets $1$ and $2$ and $8$dpi for data set $3$. The difference between data set $1$ and $2$ is that for data set $1$ banknotes have a skew up to $20^\circ$. See Figure \ref{ExampleImages} for example images.
All data sets contain 40 different banknote classes consisting of the four orientations of EUR\_005\_a, EUR\_005\_b, EUR\_010\_a, EUR\_010\_b, EUR\_020\_a, EUR\_020\_b, EUR\_050\_a, EUR\_100\_a, EUR\_200\_a and EUR\_500\_a where a and b denote the first and second series of the Euro banknotes. The different orientations are treated as different classes denoted e.g. for EUR\_005\_a by EUR\_005\_a\_1, EUR\_005\_a\_2, EUR\_005\_a\_3 and EUR\_005\_a\_4 meaning front side, front side upside down, back side and back side upside down. Furthermore category 1 images are provided which belong to neither of the trained banknote classes or are genuine Euro notes not recognized by the classical algorithm used in the field which is usually due to large folded corners. The number of images per class varies from 291 in case of EUR\_200\_a\_2 to 54091 in case of EUR\_050\_a\_3 which is due to different ratios of the denominations in the field. Per data set more than 400000 banknote images of category 4 are provided. The images contain recordings of banknotes from the field which have different fitness quality. 

\subsection{Comparison of different training conditions}

We fix the validation and test set ratio at $10\%$ per class, the training batch size at $300$ and use the Adam algorithm as optimizer. As parameters for the Adam optimizer we use the default values suggested in \cite[Section 8.5.3]{Goodfellow-et-al-2016}, i.e. a learning rate of $0.001$, an exponential decay rate for the first and second moment estimate of $0.9$ and $0.999$ respectively and a numerical stabilization constant of $10^{-8}$. 
We compare the accuracy, i.e. the ratio of correctly classified images from the test set for the different data sets and different training conditions with varying number of training images and training epochs see Table \ref{tab:ResultsData}. 
For all three data sets the results show clearly that a higher number of epochs and a larger training set lead to an improvement of the accuracy. For a smaller training set and a smaller number of epochs data set 2 achieves the best results followed by data set 1 and data set 3. For a larger number of epochs and a larger training set all data sets achieve an accuracy of $100\%$. It should be mentioned that we only consider results on notes which were recognized correctly by the currently used classifier in the field. In Subsection \ref{Subsec:Cat1} we will also compare results for Euro notes rejected by the currently used classifier.

\begin{table}[ht]
	\caption{Ratio of correctly classified test set images (accuracy) for data sets 1-3 under different training conditions.}
	\centering
$
\renewcommand{\arraystretch}{1.3}
		\begin{array}{c|c|c|c|c}
		& &\multicolumn{3}{c}{ \text{epochs}}\\
		\cline{3-5} 
		 \text{Data set }1& &1000 & 3000& 10000\\
		\cline{1-5}
		& 50& 99.814\% &99.842\% &99.852\%\\
		\cline{2-5}
		\pbox{5cm}{Training images\\ per class}& \text{up to } 3000 & 99.983\%&99.995\% &99.993\%\\
		\cline{2-5}
		& \text{all available} & 99.983\%&99.993\% &100.00\%\\			
		\end{array}$
		
		\vspace{0.2cm}
		\centering
		$\renewcommand{\arraystretch}{1.3}
		\begin{array}{c|c|c|c|c}
		& &\multicolumn{3}{c}{ \text{epochs}}\\
		\cline{3-5} 
		 \text{Data set }2& &1000 & 3000& 10000\\
		\cline{1-5}
		& 50& 99.873\% &99.947\% &99.967\%\\
		\cline{2-5}
		\pbox{5cm}{Training images\\ per class}& \text{up to } 3000 & 99.983\%&99.990\% &99.998\%\\
		\cline{2-5}
		& \text{all available} & 99.995\%&99.998\% &100.00\%\\			
		\end{array}$
		
\vspace{0.2cm}
\centering
$\renewcommand{\arraystretch}{1.3}
		\begin{array}{c|c|c|c|c}
		& &\multicolumn{3}{c}{ \text{epochs}}\\
		\cline{3-5} 
		 \text{Data set }3& &1000 & 3000& 10000\\
		\cline{1-5}
		& 50& 99.771\% &99.766\% &99.842\%\\
		\cline{2-5}
		\pbox{5cm}{Training images\\ per class}& \text{up to } 3000 & 99.969\%&99.993\% &100.00\%\\
		\cline{2-5}
		& \text{all available} & 99.967\%&99.995\% &100.00\%\\
		\end{array}	
	$
	\label{tab:ResultsData}
\end{table}

\subsection{Results for category 1 input}\label{Subsec:Cat1}

For the study of the classification of category 1 input we concentrate on data set 2. This data set consists of 25dpi images without skew and achieves the best accuracy results on images from the 40 considered banknote classes among the three data sets. In the following we turn to the problem of classification of category 1 input. By the ECB decision \cite{ECB_Decision_2012} input images which are obviously not banknotes because of a wrong image or format have to be rejected. Typical examples are double notes (images of two overlapping notes), transport errors (mostly because of jam), other currencies, cheques or genuine Euro notes with large folded corners.

\begin{figure}[ht]
\subfloat[Rejected input of type 'double note'.]{
\includegraphics[resolution=25, scale=0.25]{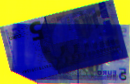}}
\quad\subfloat[Rejected input of type 'transport error'.]{
 \includegraphics[resolution=25, scale=0.25]{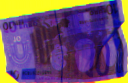}}\\
\subfloat[Rejected input of type 'other currency'.]{
  \includegraphics[resolution=25, scale=0.25]{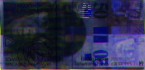}}
\quad\subfloat[Rejected input of type 'cheque'.]{
 \includegraphics[resolution=25, scale=0.25]{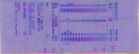}}\\
\subfloat[Rejected input of type 'rejected genuine Euro note'.]{
 \includegraphics[resolution=25, scale=0.25]{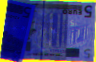}}
\caption{Examples for different types of category 1 input rejected in the field displayed at 25\% of the original size.}
	\label{Cat1_ExampleImages}
\end{figure}

In Figure \ref{Cat1_ExampleImages} we show examples for 5 different types. For the last type of 'rejected genuine Euro notes' it can be desirable to obtain a higher acceptance rate and a category 4b classification whereas a category 2 or 3 classification should be avoided.

\begin{figure}[ht]
\subfloat[Ecdfs with x-axis ranging from 0 to 1.]{
\includegraphics[trim=50 0 50 0, clip,width=8.7cm]{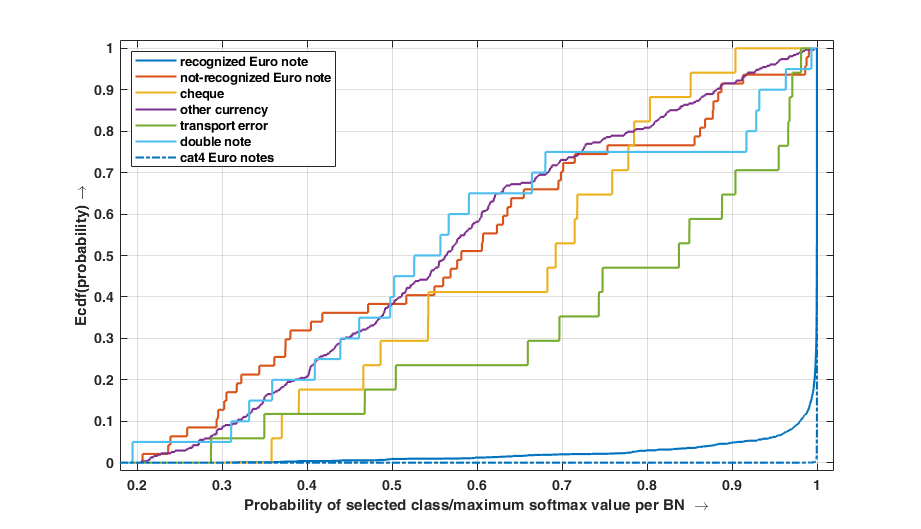}}\\
\subfloat[Ecdfs with zoom to high x-values ranging from 0.97 to 1.]{
 \includegraphics[trim=50 0 50 0, clip,width=8.7cm]{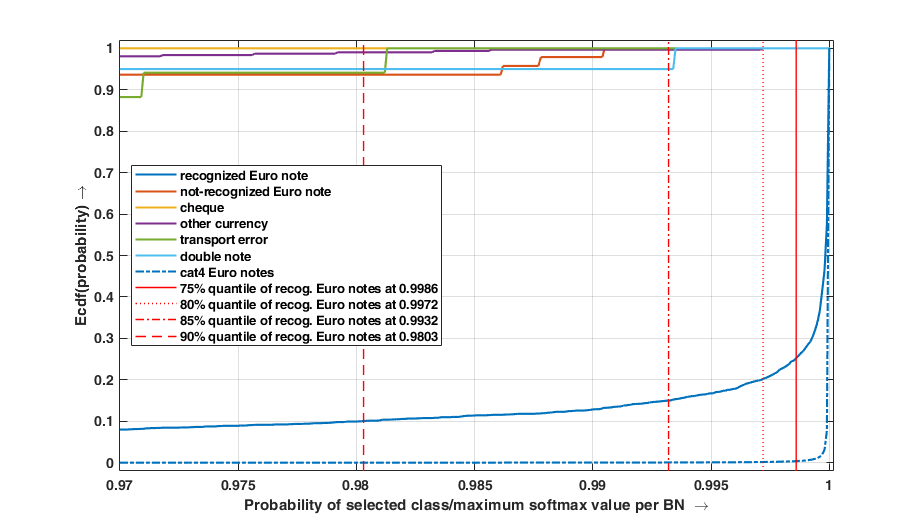}}
\caption{Empirical cumulated distribution function (ecdf) for maximal class probability in feature vector for different types of images rejected by the classical algorithm as well as for genuine notes accepted by the classical algorithm (cat 4 Euro notes).
For genuine Euro notes rejected by the classical algorithm we differ between Euro notes mapped to the correct banknote class (recognized EUR BNs) and Euro notes mapped to a wrong banknote class (not-recognized EUR BNs) by the inception-v3-classifier.} 
	\label{Cat1_Ecdfs}
\end{figure}

It should be noted that a reject rate for genuine notes below 1\% is needed to pass the ECB test procedure, see \cite{ECBTestProcedure}. So this should be the aim for a deep learning based recognition. On the other hand non-genuine category 1 images which are mapped to one of the 40 banknote classes will be considered as fake notes (category 2 or 3) in the subsequent authenticity checks. Category 2 or 3 notes have to be investigated by national central bank authorities and it is therefore not desirable that a large ratio of category 1 is sorted to category 2 or 3. Thus, a convenient threshold $T$ for the mapping to the $0$-class should not lead to a higher reject rate of genuine banknotes and should not lead to an increased amount of category 1 images mapped to a banknote class. In Table \ref{tab:SortingResultsWith0Class} we compare the reject rates for the inception-v3 net combined with a 0-class module with different thresholds (C1-C4). As test set we use 10\% of the genuine notes accepted plus 10\% of the genuine notes rejected by the classical algorithm. The thresholds for the 0-class module are motivated by the quantiles of the maximal class probabilities in the set of genuine notes rejected by the classical algorithm but recognized correctly by the inception-v3 classifier see Figure \ref{Cat1_Ecdfs}.
As shown in Table \ref{tab:SortingResultsWith0Class} reject rates of 0.51\% and lower can be obtained which meet the central bank requirement of a reject rate below 1\%. However, if the reject rate is reduced further by choosing a lower threshold for the 0-class module at some point we start to observe banknotes being sorted to the wrong banknote class (C4).

\begin{table}[ht]
\caption{Comparison of sorting results for different classifiers C1-C4.}
\renewcommand{\arraystretch}{1.5}
	\centering
$
		\begin{array}{c|c|c|c|c}
		\text{Nr.}&\text{Classifier}& \pbox{5cm}{reject rate\\ on genuine\\ BNs (\%)} & \pbox{5cm}{\# Non Euro\\ category 1\\ images\\ sorted\\ in BN class} 
		& \pbox{5cm}{\# accepted\\ genuine BNs\\in wrong\\ BN class}  \vspace{0.1cm}\\
		\hline
		 \text{C1}&\pbox{5cm}{\vspace{0.1cm}Inception-v3+\\0-Class module\\ with T=0.9986\vspace{0.1cm}}&0.51&0&0\\
		\hline
		 \text{C2}&\pbox{5cm}{\vspace{0.1cm}Inception-v3+\\0-Class module\\ with T=0.9972\vspace{0.1cm}}&0.26&1&0\\
		\hline
		 \text{C3}&\pbox{5cm}{\vspace{0.1cm}Inception-v3+\\0-Class module\\ with T=0.9932\vspace{0.1cm}}&0.12&2&0\\
			\hline
		 \text{C4}&\pbox{5cm}{\vspace{0.1cm}Inception-v3+\\0-Class module\\ with T=0.9803\vspace{0.1cm}}&0.04&5&3\\
		\end{array}	
		$
	\label{tab:SortingResultsWith0Class}
\end{table}

%

\subsection{Run time measurements}
We distinguish between training time and test time. 
The training time consists of the time for feature extraction, i.e. for the application of the pretrained inception-v3 net to the image which is displayed in Table \ref{tab:FeatureExtractionTime} and of the time for retraining of the last layer displayed in Table \ref{tab:RetrainingTime}. Under a practical view the time for retraining is neglectable whereas the feature extraction time for all images can be more than 2h.
Still this is a minor problem since this time has to be expended only once. The test time is much more critical since in practice we are facing strict real time requirements in the contemplated application. The test time is the sum of the feature extraction time for 1 image (about 20ms by column 1 of Table \ref{tab:FeatureExtractionTime}) and the test time of the resulting feature vector which is in average 0.59ms. Thus we obtain a complete test time of about 20ms for images of all three data sets. Here we should mention that the test time is measured on the same workstation as the training time which is equipped with a rather expensive graphical processing unit.
\begin{table}[h!t]
\caption{Feature extraction time for different amounts of images from data sets 1-3.}
\renewcommand{\arraystretch}{1.3}
\centering
$
		\begin{array}{c|c|c|c|c}
		& \multicolumn{4}{c}{ \text{Number of images}}\\
		\hline
		 &1&2000& 120000&400000  \\
		\cline{1-5} 
		 \text{data set } 1 &20.04\text{ms}&40.08\text{s}& 40\text{min } 48\text{s}&136\text{min}  \\
		\cline{1-5}
		 \text{data set } 2& 19.73\text{ms}&39.46\text{s} &39\text{min } 28\text{s}&131\text{min } 33\text{s} \\
		\cline{1-5}
		 \text{data set } 3& 19.49\text{ms}&38.98\text{s}& 38\text{min } 59\text{s}&130\text{min } 27\text{s}\\
		\end{array}
		$
	\label{tab:FeatureExtractionTime}
\end{table}

\begin{table}[h!t]
\caption{Time for retraining depending on the number of epochs and the batch size.}
\renewcommand{\arraystretch}{1.3}
	\centering
$
		\begin{array}{c|c|c|c|c}
		& &\multicolumn{3}{c}{ \text{epochs}}\\
		\cline{3-5} 
		 & &1000 & 3000& 10000\\
		\cline{1-5}
		& 30& 32.46\text{s} & 83.89\text{s} &264.69\text{s}\\
		\cline{2-5}
		\text{batch size}& 100 & 66.98\text{s}&180.60\text{s} &585.16\text{s}\\
		\cline{2-5}
		& 300 & 153.76\text{s}&444.54\text{s} &1471.91\text{s}\\
		\end{array}
		$
	\label{tab:RetrainingTime}
\end{table}

\section{Conclusion}\label{Sec:Con}
In this paper we studied the feasibility of recognition of Euro banknotes by deep learning. We focused on requirements from central banks including in particular the rejection of objects with wrong image or format. Thus, we introduce a 0-class module which can be concatenated to every classifier with a probability vector for the trained classes as output and allows the rejection i.e. sorting to a 0-class depending on a chosen threshold. In our experimental study we observed that the largest considered number of training images and epochs achieves 100\% accuracy on all three considered data sets (25dpi without skew correction, 25dpi with skew correction and 8dpi with skew correction). For a smaller number of training images (50 per class) and epochs (1000) we observe that data set 2 achieves the best accuracy (99.873\%), followed by data set 1 (99.814\%) and data set 3 (99.771\%). 
The study of sorting results using the 0-class module for rejection shows that reject rates of 0.51\% and lower can be obtained, which meets central bank requirements. However, if the reject rate is reduced further by choosing a lower threshold for the 0-class module at some point we observe banknotes being sorted to the wrong banknote class. Run time measurements show that test time is approximately 20ms on a work station equipped with a state of the art GPU. This is acceptable for the considered real-time application.
In summary, the deep learning approach with additional 0-class module offers the possibility for a low reject rate of genuine Euro banknotes, though it is subject of further investigation if a higher acceptance rate leads only to a higher category 4b rate or also to a higher category 3 rate where the latter is undesirable.

\section*{Acknowledgment}
The authors would like to thank Michael Flack and Armin St\"ockli  from CI Tech Sensors AG for supporting the presented feasibility study and the publication of the results. Furthermore the authors would like to express their gratitude to Rainer Stute from Diebold Nixdorf AG for the collection of the field data, to Peter Zemp from CI Tech Sensors AG for the preprocessing of the field data and helpful hints to the literature on banknote recognition, to Armin St\"ockli for his valuable comments and active support concerning the results on the classification of the category 1 images. Last but not least, we would like to express our special thanks to Markus S\"u{\ss} from Mittweida University of Applied Science for his advice concerning Python and the Tensorflow framework.
%


\bibliographystyle{IEEEtran}
\bibliography{SchulteStapsLampe2019}
%

%
%
%
%

\end{document}